\documentclass{INTERSPEECH2023}

\interspeechcameraready

\usepackage{here}
\usepackage{paralist}
\usepackage{multirow}
\usepackage{color, soul}
\usepackage[normalem]{ulem}
\usepackage{cite}

\makeatletter
\newcommand{\figcaption}[1]{\def\@captype{figure}\caption{#1}}
\newcommand{\tblcaption}[1]{\def\@captype{table}\caption{#1}}
\makeatother

\title{SpeechGLUE:\\How Well Can Self-Supervised Speech Models Capture Linguistic Knowledge?}
\name{Takanori Ashihara, Takafumi Moriya, Kohei Matsuura, Tomohiro Tanaka\\Yusuke Ijima, Taichi Asami, Marc Delcroix, Yukinori Honma}
\address{NTT Corporation, Japan}
\email{takanori.ashihara@ntt.com}

\begin{document}

\maketitle
 
\begin{abstract}
\vspace{-0.1cm}
Self-supervised learning (SSL) for speech representation has been successfully applied in various downstream tasks, such as speech and speaker recognition.
More recently, speech SSL models have also been shown to be beneficial in advancing spoken language understanding tasks, implying that the SSL models have the potential to learn not only acoustic but also linguistic information.
In this paper, we aim to clarify if speech SSL techniques can well capture linguistic knowledge.
For this purpose, we introduce SpeechGLUE, a speech version of the General Language Understanding Evaluation (GLUE) benchmark.
Since GLUE comprises a variety of natural language understanding tasks, SpeechGLUE can elucidate the degree of linguistic ability of speech SSL models.
Experiments demonstrate that speech SSL models, although inferior to text-based SSL models, perform better than baselines, suggesting that they can acquire a certain amount of general linguistic knowledge from just unlabeled speech data.
\end{abstract}
\noindent\textbf{Index Terms}: self-supervised learning, speech representation, linguistic knowledge, natural language processing

\vspace{-0.3cm}
\section{Introduction}
\vspace{-0.2cm}
Self-supervised learning (SSL) has become a prominent technique to leverage a large amount of unlabeled data in an unsupervised fashion.
For the speech community, various SSL methods have been proposed~\cite{cpc, wav2vec2, hubert, data2vec, wavlm} and accuracy has been dramatically improved, especially in automatic speech recognition (ASR) tasks under low-resource conditions.
Subsequent studies have demonstrated success with task-generalizability, i.e.,~performance improvement in a wide range of tasks such as speaker recognition, emotion recognition, and speech enhancement~\cite{superb, superb_sg}.
These positive results likely reflect the ability of the SSL model to learn a wide range of speech information (e.g.,~phonemes and speaker characteristics) from only speech data without any labels~\cite{cpc, ssl_review, ssl_piece, ashihara_is}.
Actually, a previous SSL study has demonstrated a clear relationship between latent representations and phonetic units~\cite{hubert, ssl_piece}.
\par
More recently, speech SSL models have also been utilized in spoken language understanding (SLU) tasks~\cite{espnet_slu, slue, slue_compara}, e.g.,~named entity recognition and sentiment analysis, and these universal models have demonstrated superiority over conventional approaches.
Their success can be naturally attributed to the speech information captured by SSL as described above.
However, since performing SLU tasks requires natural language processing (NLP) ability, the benefit of SSL in these tasks may also imply that speech SSL models can latently capture linguistic characteristics, such as semantics and syntax, from speech signals in addition to acoustic characteristics.
\par
The above implications are supported by other studies.
Previous studies~\cite{zero1, zero2, zero3} have presented and utilized the zero-resource benchmark to evaluate the spoken language models, which are language models trained using the discrete acoustic units obtained by clustering the speech SSL output.
Benchmark results have shown that the unit-based language models are feasible, indicating that self-supervised representation seemingly retains some multi-level information such as phonetics, lexicon, syntax, and semantics.
The other work of~\cite{ssl_word_seman, compara_layerwise} comprehensively analyzed the speech representation layer-wise and demonstrated that SSL models capture some word and semantic information in the middle layers.
While the aforementioned papers shed light on the language properties acquired by the speech SSL, further investigation is needed because, for example, it is important to align the representations across speech and text modalities for a unified multimodal SSL model~\cite{slam, mslam}.
In particular, motivated by the above efforts, we aim to elucidate if linguistic information captured by speech SSL models is enough to solve practical and diverse natural language understanding (NLU) tasks.
Moreover, we would like to compare the linguistic capabilities of not only speech SSL models but also text-based ones such as BERT~\cite{bert} and identify their main differences to confirm if text data is still required to represent the linguistic information.
\par
In this paper, for the purpose of exploiting the linguistic knowledge learned via SSL, we apply a probing task, which is a popular assessment method, to self-supervised speech representations.
Specifically, we introduce the speech version of the General Language Understanding Evaluation (GLUE) benchmark~\cite{glue}, called SpeechGLUE, and evaluate speech SSL models extensively, in a fair comparison to NLP SSL models.
While there is a large body of NLU probing tasks and benchmarks ~\cite{glue, superglue, big_bench}, we adopt GLUE which is relatively basic among the existing NLU benchmarks.\footnote{Note that our approach can be applied to any probing task of NLP.}
Since GLUE is designed to cover a diverse range of NLU tasks, SpeechGLUE, a collection of NLU tasks that convert input text to speech, is intended to evaluate the general-purpose NLU knowledge within the speech SSL models.
For the conversion, we adopt text-to-speech (TTS) systems, which allow for the realization of tasks that assess purely linguistic knowledge by constraining acoustic conditions such as variations in speakers, speaking styles and recording settings.
We base our implementation of SpeechGLUE on the S3PRL toolkit developed for SUPERB~\cite{superb}, which facilitates comparisons with various speech SSL models.\footnote{We release SpeechGLUE for reproducibility and comparison with successive SSL techniques at \url{https://github.com/ashi-ta/speechGLUE}.}
\par
From published experiments, speech SSL models lag behind NLP SSL models in performance, especially in the task of judging whether a sentence is linguistically acceptable.
However, strong speech SSL models, such as WavLM {\sc Large}, perform substantially better than chance level or baselines (e.g.,~log-mel filterbank output) and achieve close to the performance of the NLP SSL models, especially in sentence similarity and natural language inference (NLI) tasks.
The experiments confirm that SSL models can capture enough linguistic information to tackle purely NLU tasks.
By releasing the SpeechGLUE task, we hope to not only clarify the linguistic capabilities of current SSL models, but also to allow future models to be assessed in terms of their improvements in these tasks.
Indeed, we believe that SLU tasks, which require capturing fine linguistic information, will be more and more important in future speech processing research.
\vspace{-0.3cm}
\section{Related work}
\vspace{-0.2cm}
There are several speech benchmarks related to the current work.
ASR-GLUE~\cite{asr_glue} is a collection of human speech recordings based on selected sentences intended for some of the GLUE tasks.
This benchmark includes only development and test sets to evaluate the negative impact of ASR error propagating to the backend NLU system.
Therefore, we cannot train downstream models on this dataset, making it unsuitable for our purpose.
Other datasets such as~\cite{superb, noss} have been helpful in benchmarking speech SSL models.
While they are designed to evaluate the generalizability through diverse speech processing tasks, SpeechGLUE, a collection of purely NLU tasks based on GLUE, aims to delve into the linguistic properties.
\vspace{-0.4cm}
\section{Method}
\vspace{-0.2cm}
\label{sec:method}
In this section, we briefly explain GLUE~\cite{glue} tasks in Section~\ref{ssec:glue} and then, introduce SpeechGLUE in Section~\ref{ssec:speechglue}.
\vspace{-0.2cm}
\subsection{GLUE}
\vspace{-0.2cm}
\label{ssec:glue}
The original GLUE benchmark contains 9 tasks divided into 3 categories:~\begin{inparaenum}[1)]
\item the Corpus of Linguistic Acceptability (CoLA)~\cite{cola} and the Stanford Sentiment Treebank (SST-2)~\cite{sst2} for single-sentence tasks,
\item the Microsoft Research Paraphrase Corpus (MRPC)~\cite{mrpc}, Quora Question Pairs (QQP)~\cite{qqp} and the Semantic Textual Similarity Benchmark (STS-B)~\cite{stsb} for similarity and paraphrase tasks, and
\item Multi-Genre NLI (MNLI)~\cite{mnli}, Question-answering NLI (QNLI)~\cite{qnli}, Recognizing Textual Entailment (RTE)~\cite{rte1, rte2, rte3, rte5} and Winograd NLI (WNLI)~\cite{wnli} for NLI tasks.
\end{inparaenum}
An overview of each task is given in Table~\ref{tab:glue}.

\begin{table}[t]
\caption{Brief summary of GLUE. MCC, PCC, and SCC denote Matthews, Pearson, and Spearman correlation coefficients, respectively. For details on this benchmark, see the original paper of~\cite{glue}.}
\vspace{-0.3cm}
\label{tab:glue}
\centering
\scalebox{0.6}[0.6]{
\begin{tabular}{l|lll}
\toprule
Corpus  &  Task   &   Metrics   &  Labels  \\
\midrule
\multicolumn{4}{c}{\bf{Single-sentence tasks}} \\
\midrule
CoLA   &  acceptability (grammaticality)   &  MCC   &  unacceptable / acceptable \\
SST2   & sentiment analysis &  accuracy   & positive / negative  \\
\midrule
\multicolumn{4}{c}{\bf{Similarity and paraphrase tasks using sentence pairs}} \\
\midrule
MRPC   & semantic equivalence (paraphrase)  & accuracy \& F1  & equivalent / not equivalent   \\
QQP    &  semantic equivalence (paraphrase)  & accuracy \& F1  & duplicate / not duplicate  \\
STS-B   &  sentence similarity   &  PCC \& SCC  &  similarity score (1--5)  \\
\midrule
\multicolumn{4}{c}{\bf{Natural language inference (NLI) tasks using sentence pairs}} \\
\midrule
MNLI-m   & NLI (in-domain)   & \multirow{2}{*}{accuracy} & entailment / contradiction / \\
MNLI-mm    &  NLI (cross-domain)   &    &  \hspace{0.3cm}neutral  \\
QNLI    &  NLI (question-answering)   &  accuracy  &  entailment / not entailment  \\
RTE   &  NLI  &  accuracy  &  entailment / not entailment  \\
WNLI   &  NLI (coreference)  &  accuracy  &  entailment / not entailment  \\
\bottomrule
\end{tabular}
}
\vspace{-0.6cm}
\end{table}
\vspace{-0.2cm}
\subsection{SpeechGLUE}
\vspace{-0.2cm}
\label{ssec:speechglue}
As described in Section~\ref{ssec:glue}, the GLUE benchmark was originally designed to assess NLU systems.
Since our objective is to evaluate speech SSL models in terms of NLP capability, the text sentences must be converted into corresponding speech utterances.
In this work, we applied a single-speaker TTS system to investigate purely linguistic knowledge of speech SSL by suppressing acoustic variabilities such as speaker, speaking style, and recording noise.
Recent neural-based TTS systems can generate high-quality speech in terms of naturalness and intelligibility~\cite{tts_review}.
Moreover, the conversion cost via TTS is lower than rerecording the text by humans.
\par
\begin{figure}[t]
  \centering\hspace*{0.2cm}
  \includegraphics[width=6cm]{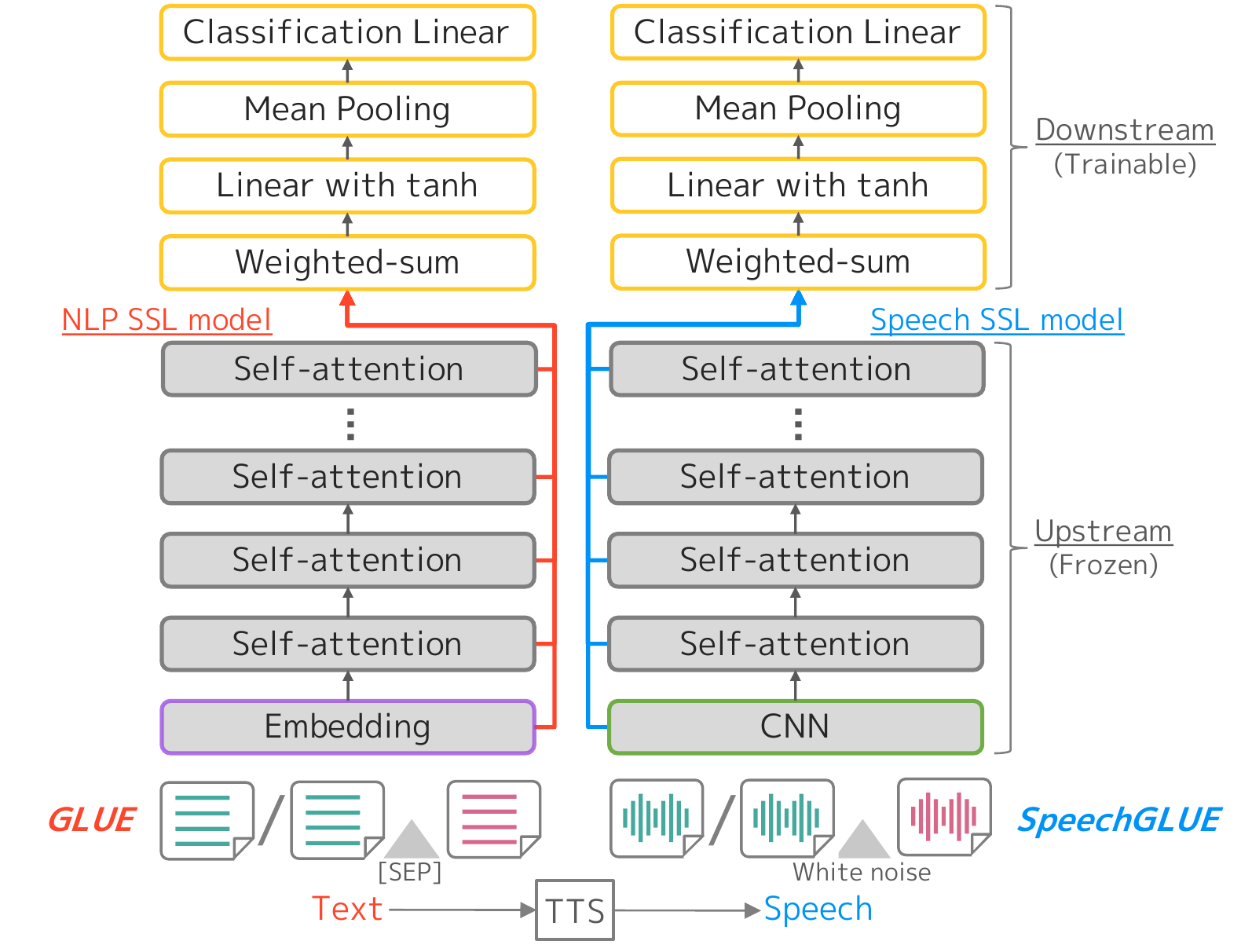}
  \vspace{-0.3cm}
  \caption{Schematic diagrams of GLUE and SpeechGLUE.}
  \label{fig:system}
\vspace{-0.6cm}
\end{figure}

The overall system is summarized in Figure~\ref{fig:system}.
Since SpeechGLUE is essentially a counterpart of GLUE, speech and NLP SSL models are fairly comparable except for the \texttt{[SEP]} token utilized in NLP SSL models.
\texttt{[SEP]} token is a special separation token that indicates where to split pairs in the examples (e.g.,~pairs of question-answer).
For speech SSL models, this study simply employs a white noise signal of 50 ms as an alternative.
\par
To incorporate SSL upstream models, we utilize the pre-trained model as a feature extractor and do not update the parameters during the training on GLUE/SpeechGLUE tasks in order to verify just the linguistic representation captured only by SSL.

\vspace{-0.4cm}
\section{Experimental Setup}
\vspace{-0.1cm}
\subsection{SpeechGLUE dataset}
\vspace{-0.2cm}
\label{exs:speechglue}
For the GLUE benchmark itself, we utilized the publicly available dataset\footnote{\url{https://huggingface.co/datasets/glue}} provided by Hugging~Face.
To generate the speech, we adopted the VITS~\cite{vits} model\footnote{\url{https://huggingface.co/espnet/kan-bayashi_ljspeech_vits}} trained by LJSpeech\footnote{\url{https://keithito.com/LJ-Speech-Dataset}} using the ESPnet toolkit~\cite{espnet}.
Because the VITS model was trained with a sampling frequency of 22050 Hz, we resampled the output data to 16000 Hz to match the sampling frequency assumed by the SSL models.
The dataset after applying TTS is summarized in Table~\ref{tab:dataset}.
Note that the original number of examples in the test set of QQP and in the training set of SST2 were 390965 and 67349, but the sizes were reduced to 390963 and 67347 through the execution of TTS.
This is because some samples were deemed impractical as they included a huge number of digits or only null text, which could not be synthesized into speech.
In addition, the original text samples were altered with the ESPnet-based text normalization such as by removing symbols (e.g.,~quotation marks) and by translating Latin abbreviations (e.g.,~``{\it i.e.}'') into English (e.g.,~{\it ``that is''}).
Thus, the word sequence of the GLUE benchmark was also transformed for a fair comparison; nevertheless, no significant degradation was noted in our preliminary GLUE experiment.
\begin{table}[t]
\caption{Summary of the data size in SpeechGLUE. The tasks with underline indicate the relatively high-resource tasks. Note that the number of examples in some tasks is decreased from the original number as noted in Section~\ref{exs:speechglue}.}
\vspace{-0.3cm}
\label{tab:dataset}
\centering
\scalebox{0.60}[0.60]{
\begin{tabular}{l|rrr}
\toprule
\multirow{2}{*}{Corpus}  & \multicolumn{3}{c}{\#hours (\#examples)}    \\ \cline{2-4}
                        & Training                     & Development   & Test  \\
\midrule
\multicolumn{4}{c}{\bf{Single-sentence tasks}} \\
\midrule
CoLA                   & 6.3 (8551)    &0.8 (1043)     & 0.8 (1063)  \\
\uline{SST2}                      & 66.5 (67347) & 1.6 (872)   & 3.4 (1821)  \\
\midrule
\multicolumn{4}{c}{\bf{Similarity and paraphrase tasks (first / second sentence)}} \\
\midrule
MRPC                       & 8.4 / 8.4 (3668)     & 0.9 / 0.9 (408)   & 3.9 / 3.9 (1725)   \\
\uline{QQP}                    & 399.3 / 404.0 (363846)       & 44.4 / 44.9 (40430)   & 438.8 / 437.4 (390963)   \\
STS-B           & 6.6 / 6.6 (5749)      & 1.9 / 1.8 (1500)   & 1.5 / 1.5 (1379)   \\
\midrule
\multicolumn{4}{c}{\bf{Natural language inference (NLI) tasks (first / second sentence)}} \\
\midrule
\uline{MNLI-m}         & \multirow{2}{*}{811.2 / 399.7 (392702)}    &  19.9 / 10.0 (9815) & 20.1 / 9.9 (9796) \\
\uline{MNLI-mm}             &        & 21.0 / 11.1 (9832) &  21.1 / 11.1 (9847) \\
\uline{QNLI}               & 111.0 / 332.9 (104743)      & 5.8 / 17.8 (5463)    & 5.8 / 18.0 (5463)   \\
RTE              & 12.7 / 2.6 (2490)      & 1.4 / 0.3 (277)   & 14.4 / 3.1 (3000)   \\
WNLI             & 1.2 / 0.5 (635)    & 0.1 / 0.1 (71)   & 0.5 / 0.2 (146)  \\
\bottomrule
\end{tabular}
}
\vspace{-0.3cm}
\end{table}

\vspace{-0.2cm}
\subsection{Upstream}
\vspace{-0.2cm}
\label{exs:upstream}
To explore the linguistic ability of speech SSL models, we utilized the multiple baselines and SSL methods with speech and text modalities summarized in Table~\ref{tab:upstreams}.
In this table, the upstream components were divided into three sections:~the baselines, the speech SSL models, and the NLP SSL models.
For the model architecture, the encoder of all SSL models with {\sc Base} ({\sc Large}) structure consisted of 12~(24) Transformer blocks with 768-dim~(1024-dim) embeddings, 3072-dim~(4096-dim) feed-forward networks, and attention heads of 12~(16).
As explained in Section~\ref{exs:speechglue}, the parameters of all SSL models were frozen during training.

\begin{table}[t]
\caption{Overview of upstream models. \texttt{LS}, \texttt{LL}, \texttt{GS}, \texttt{VP}, \texttt{MLS}, \texttt{CV}, \texttt{VL}, \texttt{BBL}, \texttt{BC} and \texttt{EW} denote LibriSpeech, Libri-Light, GigaSpeech, VoxPopuli, Multilingual LibriSpeech, CommonVoice, VoxLingua107, BABEL, BookCorpus, and English Wikipedia, respectively. Note that the \texttt{VP} utilized in WavLMs is the subset of only English data.}
\vspace{-0.3cm}
\label{tab:upstreams}
\centering
\scalebox{0.6}[0.6]{
\begin{tabular}{l|rll}
\toprule
Upstreams & \#Params & Input   &  Unlabeled data (\#hours or \#words) \\
\midrule
FBANK  & -  & waveform       & -   \\
w/o SSL {\sc Large}  & 315M  & waveform     & -  \\
Phoneme  & 0.01M   & text       & - \\
\midrule
wav2vec2.0 {\sc Base}~\cite{wav2vec2}  & 94M  & waveform       & \texttt{LS} (960 hours)   \\
wav2vec2.0 {\sc Large}~\cite{wav2vec2}  & 315M  & waveform       & \texttt{LL} (60k hours) \\
HuBERT {\sc Base}~\cite{hubert}  & 94M  & waveform     & \texttt{LS} (960 hours) \\
HuBERT {\sc Large}~\cite{hubert}  & 315M  & waveform       & \texttt{LL} (60k hours) \\
data2vec-s {\sc Base}~\cite{data2vec}  & 94M   & waveform       & \texttt{LS} (960 hours) \\
data2vec-s {\sc Large}~\cite{data2vec}  & 315M & waveform       & \texttt{LL} (60k hours) \\
WavLM {\sc Base}~\cite{wavlm}  & 94M   & waveform       & \texttt{LS} (960 hours) \\
WavLM {\sc Base+}~\cite{wavlm}  & 94M   & waveform       & \texttt{LL} + \texttt{GS} + \texttt{VP} (94k hours) \\
WavLM {\sc Large}~\cite{wavlm}  & 315M  & waveform       & \texttt{LL} + \texttt{GS} + \texttt{VP} (94k hours) \\
XLS-R (0.3B)~\cite{xlsr}  & 315M  & waveform       & \texttt{VP} + \texttt{MLS} + \texttt{CV} + \texttt{VL} + \texttt{BBL} (436k hours)  \\
\midrule
data2vec-t {\sc Base}~\cite{data2vec}  & 125M   & text       & \texttt{BC} + \texttt{EW} (3300M words)    \\
BERT {\sc Base}~\cite{bert}  & 110M  & text       & \texttt{BC} + \texttt{EW} (3300M words) \\
BERT {\sc Large}~\cite{bert}  & 340M   & text       & \texttt{BC} + \texttt{EW} (3300M words)   \\
\bottomrule
\end{tabular}
}
\vspace{-0.7cm}
\end{table}
\par
For the baseline models listed in the first section in Table~\ref{tab:upstreams}, we adopted three types of upstream feature extractors.
FBANK was the 80-dimensional log-mel filterbank output combined with delta and delta-delta features.
We also evaluated a randomly initialized model with {\sc Large} architecture (i.e.,~w/o SSL {\sc Large}; see the second row of Table~\ref{tab:upstreams}).
We adopted this model to test the extent to which SpeechGLUE could be accurately handled by the structure itself since the Transformer architecture can inherently access long-range context.
This model had input of raw waveforms, and hence, a feature encoder with subsampling was added before the Transformer blocks.
The architecture of the feature encoder was identical to that of an existing SSL study~\cite{wav2vec2} and comprised a 7-layer convolutional neural network (CNN).
The third baseline was grapheme-to-phoneme (G2P) conversion followed by a 128-dim embedding layer to investigate the performance of ideal speech units without higher-level context.
In this paper, since we utilized ESPnet for the TTS system as described in Section~\ref{exs:speechglue}, the converter was in accordance with the G2P\footnote{\url{https://github.com/Kyubyong/g2p}} utilized inside ESPnet.
\par
As the speech SSL models, we evaluated publicly-available models with a combination of four SSL approaches and two model sizes as shown in the second section in Table~\ref{tab:upstreams}.
Specifically, we employed wav2vec2.0~\cite{wav2vec2}, HuBERT~\cite{hubert}, data2vec-s~\cite{data2vec} and WavLM~\cite{wavlm} for the SSL method, and {\sc Base} and {\sc Large} for the model size.
These models passed raw waveforms to a 7-layer CNN before the Transformer encoder as the baseline model of w/o SSL {\sc Large}.
Note that data2vec-s and data2vec-t, described below, were the speech and NLP versions of data2vec, respectively.
\par
We evaluated not only speech SSL models but also NLP SSL models in this paper.
Since the NLP SSL models were specialized to obtain language representation from large unlabeled texts, we treated the results as the performance upper bound in SpeechGLUE tasks.
The vocabulary size of data2vec-t and BERT were 50265 and 30522 subwords, resulting in the number of parameters being different.
For BERT models, segment embedding, which encodes which of the two sentences contains the subword, was always set to zero; because the embedding was not used by the speech SSL models and data2vec-t, and was disabled for a fair comparison.
In addition, recent research~\cite{segment_emb} and our preliminary experiment reported no significant difference in performance with and without the embedding.
When evaluating the NLP SSL models, we utilized the modified version of the GLUE benchmark due to text normalization for TTS as explained in Section~\ref{exs:speechglue}.
Moreover, the normalized text was composed of lowercase, and hence, we utilized the uncased BERT models.\footnote{\url{https://huggingface.co/bert-base-uncased} and \url{https://huggingface.co/bert-large-uncased} for {\sc Base} and {\sc Large} architecture. Note that, for data2vec-t, we used the only publicly available case-sensitive model at \url{https://huggingface.co/facebook/data2vec-text-base}.}

\begin{table*}[ht]
\caption{Evaluation result for each model and each task on the development set of SpeechGLUE and GLUE. Acc, MCC, PCC and SCC denote accuracy, Matthews, Pearson and Spearman correlation coefficients, respectively. The resulting score with bold font~(underline) indicates the highest score among the speech~(NLP) SSL models.}
\vspace{-0.3cm}
\label{tab:result}
\centering
\scalebox{0.70}[0.70]{
\begin{tabular}{c|c|cc|ccc|cccc|c}
\toprule
\multirow{3}{*}{Upstream group} & \multirow{3}{*}{Upstream model}    & \multicolumn{2}{c|}{\bf{Single sentence}} & \multicolumn{3}{c|}{\bf{Similarity and paraphrase}}  & \multicolumn{4}{c|}{\bf{Natural language inference (NLI)}}  &  Avg.   \\ \cline{3-11}
&    & \multicolumn{1}{c|}{CoLA}   & SST-2  & \multicolumn{1}{c|}{MRPC}    & \multicolumn{1}{c|}{QQP}      & STS-B   & \multicolumn{1}{c|}{MNLI-m / -mm}    & \multicolumn{1}{c|}{QNLI} & \multicolumn{1}{c|}{RTE} & WNLI & w/o    \\ \cline{3-11}
&    & \multicolumn{1}{c|}{MCC}    & Acc    & \multicolumn{1}{c|}{Acc (F1)} & \multicolumn{1}{c|}{Acc (F1)} & PCC (SCC) & \multicolumn{1}{c|}{Acc} & \multicolumn{1}{c|}{Acc}  & \multicolumn{1}{c|}{Acc} & Acc  & WNLI  \\
\hline
\midrule
\multirow{4}{*}{Baselines}  & Chance rate & \multicolumn{1}{c|}{-} & \multicolumn{1}{c|}{50.9}  & \multicolumn{1}{c|}{68.4 (81.2)}  & \multicolumn{1}{c|}{63.2 (0.0)}  & \multicolumn{1}{c|}{-} & \multicolumn{1}{c|}{35.4 / 35.2}  & \multicolumn{1}{c|}{50.5}  & \multicolumn{1}{c|}{52.7}  & \multicolumn{1}{c|}{56.3} & - \\ 
& FBANK & \multicolumn{1}{c|}{3.3}  & \multicolumn{1}{c|}{64.6}  & \multicolumn{1}{c|}{70.8 (81.8)}  & \multicolumn{1}{c|}{70.8 (53.8)}  & \multicolumn{1}{c|}{12.4 (10.0)}  & \multicolumn{1}{c|}{39.9 / 40.4}  & \multicolumn{1}{c|}{57.9}  & \multicolumn{1}{c|}{52.7}  & \multicolumn{1}{c|}{43.7}  & 45.9 \\ 
& w/o SSL {\sc Large} & \multicolumn{1}{c|}{6.6}  & \multicolumn{1}{c|}{55.0}  & \multicolumn{1}{c|}{68.4 (81.2)}  & \multicolumn{1}{c|}{64.1 (20.4)}  & \multicolumn{1}{c|}{8.5 (7.7)}  & \multicolumn{1}{c|}{35.1 / 35.0}  & \multicolumn{1}{c|}{54.9}  & \multicolumn{1}{c|}{53.4}  & \multicolumn{1}{c|}{62.0}  & 42.3 \\ 
& Phoneme & \multicolumn{1}{c|}{0.0}  & \multicolumn{1}{c|}{62.0}  & \multicolumn{1}{c|}{71.8 (82.4)}  & \multicolumn{1}{c|}{65.0 (35.5)}  & \multicolumn{1}{c|}{15.4 (14.8)}  & \multicolumn{1}{c|}{37.6 / 37.1}  & \multicolumn{1}{c|}{58.3}  & \multicolumn{1}{c|}{57.8}  & \multicolumn{1}{c|}{42.3}  & 45.0 \\ 
\midrule
\multirow{10}{*}{Speech SSL} & wav2vec2.0 {\sc Base}~\cite{wav2vec2} & \multicolumn{1}{c|}{5.5}  & \multicolumn{1}{c|}{65.8}  & \multicolumn{1}{c|}{71.6 (81.2)}  & \multicolumn{1}{c|}{71.2 (58.7)}  & \multicolumn{1}{c|}{45.8 (45.5)}  & \multicolumn{1}{c|}{42.9 / 43.7}  & \multicolumn{1}{c|}{63.7}  & \multicolumn{1}{c|}{56.7}  & \multicolumn{1}{c|}{36.6}  & 51.9 \\ 
& wav2vec2.0 {\sc Large}~\cite{wav2vec2} & \multicolumn{1}{c|}{0.0}  & \multicolumn{1}{c|}{73.3}  & \multicolumn{1}{c|}{72.5 (81.8)}  & \multicolumn{1}{c|}{76.1 (67.1)}  & \multicolumn{1}{c|}{58.1 (58.1)}  & \multicolumn{1}{c|}{47.4 / 49.3}  & \multicolumn{1}{c|}{73.8}  & \multicolumn{1}{c|}{56.0}  & \multicolumn{1}{c|}{\bf{57.7}}  & 56.3 \\ 
& HuBERT {\sc Base}~\cite{hubert} & \multicolumn{1}{c|}{3.1}  & \multicolumn{1}{c|}{73.3}  & \multicolumn{1}{c|}{70.3 (81.4)}  & \multicolumn{1}{c|}{72.3 (60.5)}  & \multicolumn{1}{c|}{50.4 (50.9)}  & \multicolumn{1}{c|}{44.8 / 46.1}  & \multicolumn{1}{c|}{64.3}  & \multicolumn{1}{c|}{\bf{57.4}}  & \multicolumn{1}{c|}{36.6}  & 53.6 \\ 
& HuBERT {\sc Large}~\cite{hubert} & \multicolumn{1}{c|}{24.8}  & \multicolumn{1}{c|}{81.0}  & \multicolumn{1}{c|}{73.0 (82.1)}  & \multicolumn{1}{c|}{81.0 (72.8)}  & \multicolumn{1}{c|}{70.0 (70.5)}  & \multicolumn{1}{c|}{60.7 / 62.7}  & \multicolumn{1}{c|}{76.3}  & \multicolumn{1}{c|}{54.9}  & \multicolumn{1}{c|}{35.2}  & 64.9 \\ 
& data2vec-s {\sc Base}~\cite{data2vec} & \multicolumn{1}{c|}{13.1}  & \multicolumn{1}{c|}{72.8}  & \multicolumn{1}{c|}{71.8 (81.7)}  & \multicolumn{1}{c|}{73.8 (62.2)}  & \multicolumn{1}{c|}{56.4 (56.8)}  & \multicolumn{1}{c|}{46.9 / 48.5}  & \multicolumn{1}{c|}{67.9}  & \multicolumn{1}{c|}{54.9}  & \multicolumn{1}{c|}{28.2}  & 56.2 \\ 
& data2vec-s {\sc Large}~\cite{data2vec} & \multicolumn{1}{c|}{20.4}  & \multicolumn{1}{c|}{77.8}  & \multicolumn{1}{c|}{73.8 (83.3)}  & \multicolumn{1}{c|}{74.6 (65.1)}  & \multicolumn{1}{c|}{59.1 (59.2)}  & \multicolumn{1}{c|}{53.5 / 55.2}  & \multicolumn{1}{c|}{71.6}  & \multicolumn{1}{c|}{54.2}  & \multicolumn{1}{c|}{47.9}  & 60.0 \\ 
& WavLM {\sc Base}~\cite{wavlm} & \multicolumn{1}{c|}{5.8}  & \multicolumn{1}{c|}{71.6}  & \multicolumn{1}{c|}{71.8 (81.6)}  & \multicolumn{1}{c|}{73.3 (63.3)}  & \multicolumn{1}{c|}{69.4 (69.8)}  & \multicolumn{1}{c|}{47.6 / 48.6}  & \multicolumn{1}{c|}{71.0}  & \multicolumn{1}{c|}{\bf{57.4}}  & \multicolumn{1}{c|}{38.0}  & 57.4 \\ 
& WavLM {\sc Base+}~\cite{wavlm} & \multicolumn{1}{c|}{6.9}  & \multicolumn{1}{c|}{74.5}  & \multicolumn{1}{c|}{72.8 (79.9)}  & \multicolumn{1}{c|}{75.3 (66.1)}  & \multicolumn{1}{c|}{74.3 (74.5)}  & \multicolumn{1}{c|}{49.2 / 50.0}  & \multicolumn{1}{c|}{73.8}  & \multicolumn{1}{c|}{54.5}  & \multicolumn{1}{c|}{40.8}  & 59.0 \\ 
& WavLM {\sc Large}~\cite{wavlm} & \multicolumn{1}{c|}{\bf{29.6}}  & \multicolumn{1}{c|}{\bf{82.7}}  & \multicolumn{1}{c|}{\bf{75.7 (83.0)}}  & \multicolumn{1}{c|}{\bf{83.3 (76.8)}}  & \multicolumn{1}{c|}{\bf{79.5 (79.7)}}  & \multicolumn{1}{c|}{\bf{63.8 / 65.5}}  & \multicolumn{1}{c|}{\bf{80.6}}  & \multicolumn{1}{c|}{52.0}  & \multicolumn{1}{c|}{35.2}  & \bf{68.1} \\ 
& XLS-R (0.3B)~\cite{xlsr} & \multicolumn{1}{c|}{7.2}  & \multicolumn{1}{c|}{74.5}  & \multicolumn{1}{c|}{71.1 (81.2)}  & \multicolumn{1}{c|}{78.6 (69.8)}  & \multicolumn{1}{c|}{69.1 (69.2)}  & \multicolumn{1}{c|}{54.5 / 55.9}  & \multicolumn{1}{c|}{74.3}  & \multicolumn{1}{c|}{56.0}  & \multicolumn{1}{c|}{54.9}  & 60.1 \\ 
\midrule
\multirow{3}{*}{NLP SSL} & data2vec-t {\sc Base}~\cite{data2vec} & \multicolumn{1}{c|}{41.0}  & \multicolumn{1}{c|}{86.9}  & \multicolumn{1}{c|}{\uline{79.2 (84.8)}}  & \multicolumn{1}{c|}{82.2 (76.7)}  & \multicolumn{1}{c|}{80.0 (80.2)}  & \multicolumn{1}{c|}{68.4 / 69.8}  & \multicolumn{1}{c|}{83.7}  & \multicolumn{1}{c|}{\uline{58.5}}  & \multicolumn{1}{c|}{\uline{23.9}}  & 72.2 \\ 
& BERT {\sc Base}~\cite{bert} & \multicolumn{1}{c|}{49.0}  & \multicolumn{1}{c|}{\uline{90.5}}  & \multicolumn{1}{c|}{77.2 (84.2)}  & \multicolumn{1}{c|}{\uline{85.4 (80.3)}}  & \multicolumn{1}{c|}{82.8 (82.9)}  & \multicolumn{1}{c|}{69.1 / 70.2}  & \multicolumn{1}{c|}{84.4}  & \multicolumn{1}{c|}{53.1}  & \multicolumn{1}{c|}{15.5}  & 73.5 \\ 
& BERT {\sc Large}~\cite{bert} & \multicolumn{1}{c|}{\uline{51.4}}  & \multicolumn{1}{c|}{90.3}  & \multicolumn{1}{c|}{76.7 (83.6)}  & \multicolumn{1}{c|}{85.2 (80.5)}  & \multicolumn{1}{c|}{\uline{82.8 (83.1)}}  & \multicolumn{1}{c|}{\uline{70.4 / 71.0}}  & \multicolumn{1}{c|}{\uline{85.0}}  & \multicolumn{1}{c|}{53.4}  & \multicolumn{1}{c|}{14.1}  & \uline{74.0} \\ 
\bottomrule
\end{tabular}
}
\vspace{-0.6cm}
\end{table*}

\vspace{-0.3cm}
\subsection{Downstream}
\vspace{-0.2cm}
To benchmark the upstream models on the SpeechGLUE tasks, the downstream model was straightforwardly connected to the backend of the upstream models.
The parameters of downstream models were updated during training unlike upstream models, which acted as feature extractors.
The architecture of the downstream model, as motivated by an existing study~\cite{superb}, consisted of the weighted-sum of all hidden layers of upstream models followed by a 256-dim linear layer with tanh function, mean-pooling across whole sequences, and a final linear layer for classification or regression task.
With respect to the last linear layer, the number of classes was 2 except for MNLI and STS-B,  3 for MNLI, and 1 for STS-B to perform the regression task.
We applied a dropout with probability of 0.1 to the output of tanh function.
The downstream model structure is basically identical regardless of the upstream model and tasks, while the weighted-sum was not utilized for upstream models without multiple layers (i.e.,~FBANK and phoneme).
\par
For optimization, we adopted Adam with a learning rate of $3\times10^{-4}$ and a batch size of 32.
Two types of total training steps were used depending on the amount of training data:~50k steps for low-resource tasks and 150k steps for high-resource tasks, i.e.,~tasks underlined in Table~\ref{tab:dataset}.
For the loss function, the models were updated using cross-entropy loss for all tasks except STS-B which used a mean squared error loss.
\par
The entire system was evaluated on the development set of low-resource~(high-resource) tasks for every 1k~(12.5k) steps, and only the highest performances are reported here.
Note that the evaluations were conducted only on the development set and not on the private test set proceeding on the GLUE server since the goal of this study was to investigate whether or not linguistic information was acquired rather than to achieve state-of-the-art performance on the GLUE benchmark.
Additionally, there were seemingly no clear differences in performance tendency across tasks between the development and test sets from the previous NLP studies such as~\cite{cocolm}.
\par
To validate that the synthesized speech was generated properly, speech SSL models further addressed ASR tasks by using the SpeechGLUE dataset.
The training setup for the ASR task was the same as the settings of SUPERB~\cite{superb} except for low-resource tasks.
For low-resource tasks, i.e.,~tasks with no underline in Table~\ref{tab:dataset}, the total training steps were reduced to 50k and the evaluation by the development set was performed every 500 steps, in addition to changing the learning rate to $2\times10^{-4}$.
With respect to high-resource tasks, we randomly selected a maximum of 100 hours of data from the training set as in SUPERB, where only {\it train-clean-100} from LibriSpeech was utilized.
In line with this, we also randomly selected the development set to be a maximum of 5 hours and reported the best word error rates~(WERs) on the development set only as well as for the setting noted above.
From the ASR experiments, we confirmed that the average WER for all tasks except for WNLI\footnote{Since the training set in WNLI contains only 635 samples, the WER was unstable.} ranged from 13.1\% to 18.2\%.
For example, WavLM {\sc Large}, which yielded the best averaged score, was able to achieve WERs of less than 10\% on most tasks.

\vspace{-0.4cm}
\section{Results}
\vspace{-0.2cm}
The experimental results of SpeechGLUE and GLUE are shown in Table~\ref{tab:result}.
Note that, in calculating the average score, we excluded the score of the WNLI task due to its performance instability\footnote{In our preliminary experiment with wav2vec2 {\sc Base}, changing the random seed resulted in a standard deviation of 4.8\% in accuracy over the five trials, and the training itself was also unstable.} caused by the extremely small number of examples as was also pointed out in a previous NLP paper~\cite{bert}.
From the results, we can find the overall performance tendency that the NLP SSL models attain the highest performance, followed by the speech SSL models, and finally the baselines, especially in the CoLA task, which requires grammatical knowledge mainly.
Among speech SSL, as with the ASR task, WavLM {\sc Large} demonstrated the best performance.
Especially in some sentence similarity and NLI tasks, the performance was comparable to that of the NLP SSL models, suggesting that the speech SSL models can learn enough linguistic information to handle those tasks.
By comparing WavLM {\sc Base}, {\sc Base+} and {\sc Large}, we observe that model capacity is more critical than the size of unlabeled data.
However, XLS-R (0.3B), which is a multilingual model, almost matched the accuracy of WavLM {\sc Base+} despite its larger data volume and size, indicating the language dependency of the SSL model~\cite{lang_depend}.
The performance degradation of NLP SSL models in GLUE compared to the scores presented in the previous papers~\cite{data2vec, bert} may be due to the limitation that the entire model was not fine-tuned to ensure a fair comparison.
\par
\begin{figure}[t]
  \vspace{-0.1cm}
  \centering
  \includegraphics[width=5.9cm]{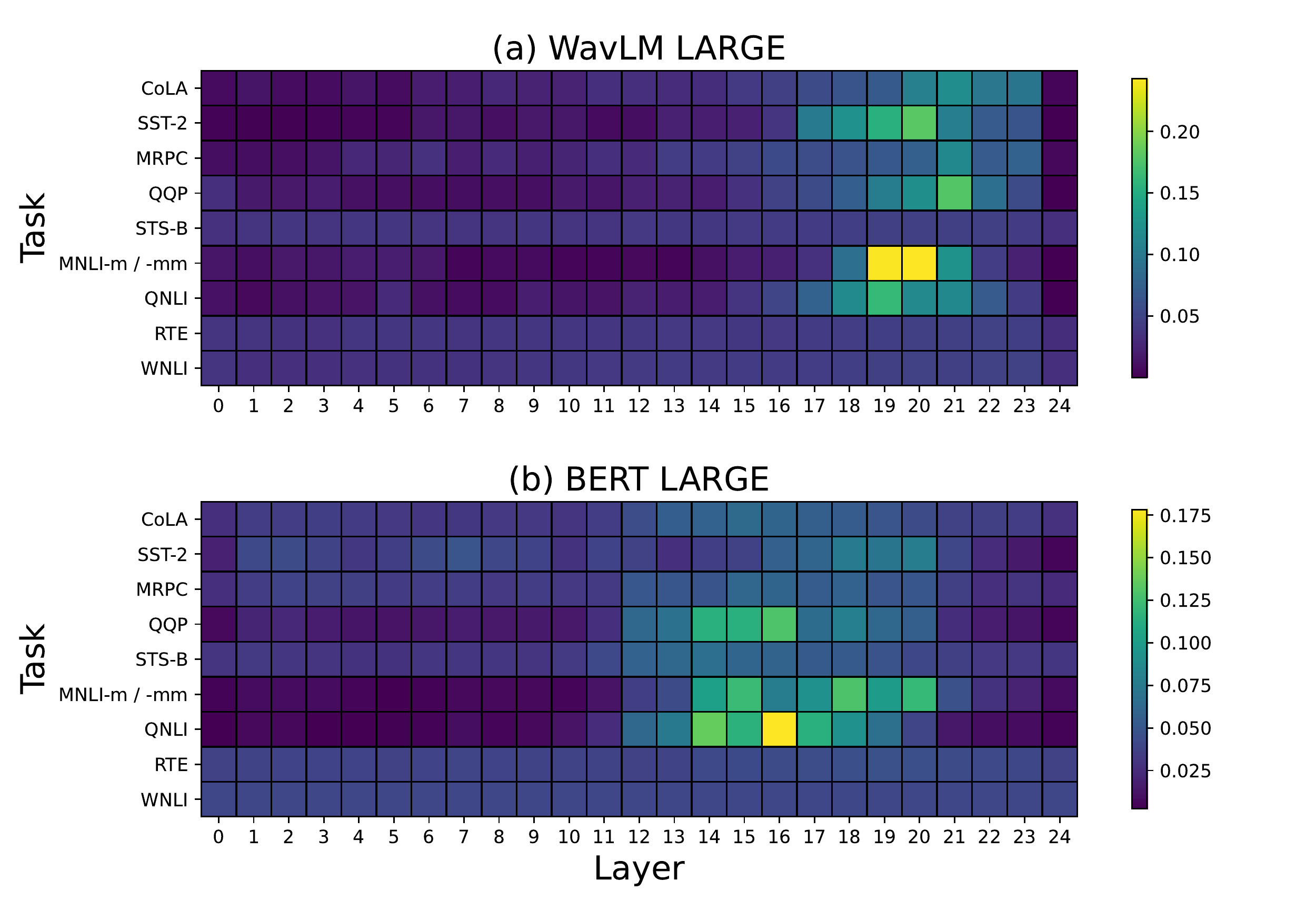}
  \vspace{-0.4cm}
  \caption{The weights of weighted-sum. The 0th layer corresponds to the input to the 1st layer of the encoder.}
  \label{fig:weight}
\vspace{-0.7cm}
\end{figure}

To investigate the contribution of features in each layer, Figure~\ref{fig:weight} depicts the weights of weighted-sum for WavLM {\sc Large} and BERT {\sc Large} on each task.
The weights of WavLM are concentrated on the layers between 18 and 24, and the features in those latter layers seem to be important in performing non-speaker-related tasks~\cite{wavlm, compara_layerwise}.
Compared with BERT, those layers are seemingly shifted somewhat later.
It is noteworthy that it seems difficult to learn clear weights for some low-resource tasks, even with an NLP SSL model.
Moreover, some tasks exploit information captured in multiple layers (e.g.,~SST-2 for WavLM), while others focus more on some layers (e.g.,~MNLI for WavLM and QNLI for BERT).
However, this tendency is not consistent for WavLM and BERT.
This seems to indicate that the layers capture very different information.
\vspace{-0.4cm}
\section{Conclusions}
\vspace{-0.2cm}
In this paper, we endeavored to uncover to what extent SSL models could capture language information through our probing tasks called SpeechGLUE.
The speech SSL models performed better than chance and baselines, indicating that the pre-trained models capture some general linguistic knowledge from speech alone.
However, compared to the top-line NLP SSL models, the performance is somewhat poor in some tasks and there seems to be room for improvement through, for example, unified speech-text SSLs.
Future works contain further probing by other NLP benchmarks to analyze linguistic properties in more detail.

\clearpage
\newcommand{\BIBdecl}{\setlength{\itemsep}{-0.372 em}}
\bibliographystyle{IEEEtran}
\bibliography{mybib}

\end{document}